\pdfoutput=1

\documentclass[11pt]{article}

\usepackage[final]{acl}

\usepackage{times}
\usepackage{latexsym}

\usepackage[T1]{fontenc}

\usepackage[utf8]{inputenc}

\usepackage{microtype}

\usepackage{inconsolata}

\usepackage{graphicx}

\usepackage{amsmath} 
\usepackage{amssymb}
\usepackage{amsfonts}
\usepackage{booktabs} 
\usepackage{multirow}
\usepackage{xcolor}
\usepackage[ruled,vlined]{algorithm2e}
\usepackage{algpseudocode}
\usepackage{soul}
\usepackage{cleveref}

\definecolor{mintleaf}{RGB}{0, 184, 148}
\definecolor{dm-blue-500}{RGB}{0, 69, 177}
\definecolor{dm-purple-500}{RGB}{105,50,230}
\definecolor{mysilver}{RGB}{128,129,128}
\definecolor{my_green}{RGB}{0, 176, 80}
\definecolor{my_yellow}{RGB}{255,165,0}
\definecolor{my_red}{RGB}{255, 0, 0}
\definecolor{my_purple}{RGB}{126, 100, 158}
\definecolor{my_blue}{RGB}{49, 133, 155}
\definecolor{case_purple}{RGB}{160, 43, 147}
\definecolor{case_blue}{RGB}{15, 158, 213}

\usepackage{pifont}
\usepackage{bm} 
\newcommand{\cmark}{\textcolor{my_green}{\ding{51}}} 
\newcommand{\xmark}{\textcolor{my_red}{\ding{55}}} 

\newcommand \footnoteONLYtext[1]
{
	\let \mybackup \thefootnote
	\let \thefootnote \relax
	\footnotetext{#1}
	\let \thefootnote \mybackup
	\let \mybackup \imareallyundefinedcommand
}

\usepackage{arydshln}

\title{E$^2$CL: Exploration-based Error Correction Learning \\ for Embodied Agents}

\author{Hanlin Wang*, ~ Chak Tou Leong*, ~ Jian Wang, ~ Wenjie Li\\
  Department of Computing, The Hong Kong Polytechnic University\\
  \tt \{hanlin-henry.wang,chak-tou.leong,jian-dylan.wang\}@connect.polyu.hk\\
  \tt cswjli@comp.polyu.edu.hk
}

\begin{document}
\maketitle

\begin{abstract}

Language models are exhibiting increasing capability in knowledge utilization and reasoning. However, when applied as agents in embodied environments, they often suffer from misalignment between their intrinsic knowledge and environmental knowledge, leading to infeasible actions. Traditional environment alignment methods, such as supervised learning on expert trajectories and reinforcement learning, encounter limitations in covering environmental knowledge and achieving efficient convergence, respectively. Inspired by human learning, we propose Exploration-based Error Correction Learning (E$^2$CL), a novel framework that leverages exploration-induced errors and environmental feedback to enhance environment alignment for embodied agents. E$^2$CL incorporates \emph{teacher-guided} and \emph{teacher-free} explorations to gather environmental feedback and correct erroneous actions. The agent learns to provide feedback and self-correct, thereby enhancing its adaptability to target environments. Extensive experiments in the VirtualHome environment demonstrate that E$^2$CL-trained agents outperform those trained by baseline methods and exhibit superior self-correction capabilities.

\end{abstract}

\footnoteONLYtext{*Equal contribution.}

\section{Introduction}

Language Models (LMs) are becoming increasingly capable of knowledge utilization and reasoning across various knowledge-intensive tasks~\citep{yao2022react,lewkowycz2022solving,hao2023reasoning}. This success motivates researchers to build LM-based agents in embodied environments, which similarly requires the use of reasoning and planning upon environmental knowledge~\citep{li2022pre,xiang2024language}. 
In this case, LM-based agents are asked to plan appropriate actions based on the given environmental information and the history of actions already taken.
However, the knowledge acquired by these LM-based agents comes from general-purpose corpora during pre-training, and as a result the intrinsic knowledge of these models often \emph{misalign} with environmental knowledge.
Such environmental knowledge involves physical constraints that LMs have not yet explored.
For example, if the embodied agent holds two objects, it is prohibited to grab one more other object.
This \emph{misalignment} causes LM-based agents to frequently produce actions that cannot be executed in the environment, hindering their applications in real-world environments.

\begin{figure}[t]
  \includegraphics[width=\columnwidth]{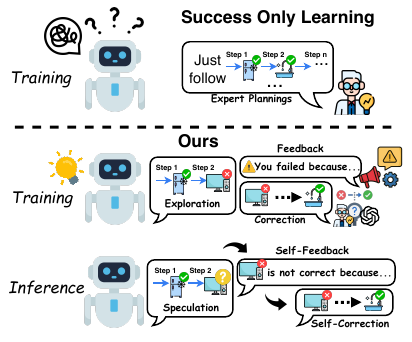}
  \caption{Traditional ``success only learning'' relies on imitating provided expert behaviors, limiting comprehensiveness. Our proposed exploration-based error correction learning (E$^2$CL) framework enhances learning by incorporating exploration-induced errors and environmental feedback during training, leading to better alignment with target environments. During inference, the agent utilizes the learned abilities to conduct self-feedback for continuous self-correction.}
  \label{fig:intro}
\end{figure}

To address the above issue, two primary types of environment alignment methods have been explored. 
The first type involves having LM-based agents undergo supervised learning on expert trajectories~\citep{li2022pre,chen2023fireact}, which are human-labeled sequences of observations and actions.
Nevertheless, these trajectories often fail to fully cover the knowledge within the environment, such as scenarios where certain actions cannot be executed. 
The second type is based on reinforcement learning (RL), which allows agents to freely explore the environment, collect trajectories that comprehensively cover the environment's knowledge, and obtain rewards based on these trajectories' success or failure~\citep{tan2024true,carta2023grounding}. 
However, the rewards are sparsely obtained because the performance evaluation of the agent is based on a complete trajectory. This makes the learning process difficult to converge.

Human learning is not comprehensive nor efficient if it relies solely on imitating experts' behavior or merely knowing whether an action is correct. Instead, by collecting and understanding feedback from the environment via exploration and learning to correct errors based on the feedback, humans can learn comprehensively and efficiently.
Inspired by this, we propose a novel exploration framework for LM-based agents to align with environments, which is called \textbf{Exploration-based Error Correction Learning (E$^2$CL)}.
As depicted in \Cref{fig:intro}, our framework incorporates exploration-induced errors and environmental feedback, leading to a comprehensive alignment with target environments.

In detail, we adopt a pretrained model as the agent to perform predefined tasks and explore the environment to collect experiences in both efficient and comprehensive manners. This is achieved by two different proposed schemes, namely \emph{teacher-guided exploration} and \emph{teacher-free exploration}.
The former prompts the agent to perform one-step exploration given sliced expert trajectories, whereas the latter allows the agent to continue exploring until it infers a stop.
In these two exploration phases, we collect the feedback given by the environment when the agent makes errors, as well as the correct actions corresponding to these error actions.
Having these exploration trajectories with additional correction, we train the agent to provide feedback on their trajectories and correct their error actions based on the feedback.
To utilize the learned self-correction ability, we further propose a \emph{speculative inference} algorithm, which performs corrections if the initially planned actions are inferred to be errors according to the feedback from the agent. 

We evaluate the agent trained by E$^2$CL in VirtualHome~(\citealp{puig2018virtualhome}), a household embodied environment. E$^2$CL-trained agent surpasses the agents trained by other baseline methods in all agentic metrics, demonstrating its effectiveness. 
Furthermore, our analysis reveals that the small models constructed using our method outperform larger models of the same series that have only undergone behavior cloning. In addition, in evaluations based on feedback-driven re-planning, our models demonstrate self-correction capabilities that are comparable to LLMs.

In summary, our main contributions are as follows. (1) We introduce the Exploration-based Error Correction Learning (E$^2$CL) framework, enabling LM-based agents to align with environments through effective feedback-driven exploration and correction. (2) We propose two novel exploration schemes, teacher-guided and teacher-free explorations, which facilitate the collection of correction and feedback via agent-environment interactions. (3) We introduce a novel action inference algorithm called speculative inference, which effectively avoids executable errors. (4) We demonstrate the superior performance of E$^2$CL-trained agents in the VirtualHome environment, surpassing baseline methods and showcasing the potential of our approach for real-world deployment. 

\section{Method}

\begin{figure*}[t]
\begin{center}
  \includegraphics[width=\linewidth]{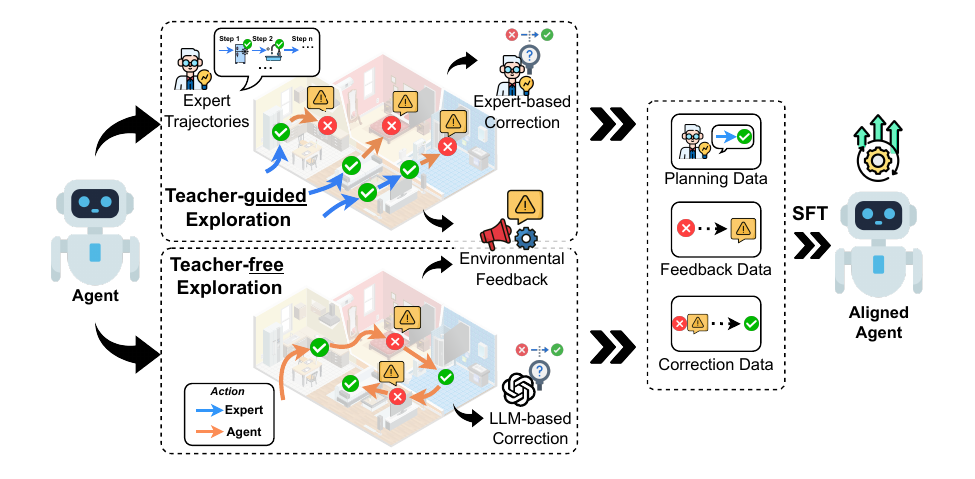}
  \caption{Overview of the proposed \textbf{Exploration-based Error Correction Learning (E$^2$CL)} framework. }
  \label{fig:framework}
\end{center}
\end{figure*}

In this section, we propose an exploration-based error correction learning (E$^2$CL) framework. This framework focuses on equipping LM-based agents with self-feedback and self-correction capabilities. The overview of  E$^2$CL is depicted in Figure \ref{fig:framework}.

\subsection{Task Formulation}

The LM-based embodied agent is asked to complete a set of tasks via interacting with a virtual environment.
The interaction between the agent and the environment can be formalized as a partially observable Markov decision process (POMDP) $\left(\mathcal{Q}, \mathcal{S}, \mathcal{A}, \mathcal{O}, \mathcal{T}, \mathcal{R}\right)$ with instruction space $\mathcal{Q}$, state space $\mathcal{S}$, action space $\mathcal{A}$, observation space $\mathcal{O}$, transition function $\mathcal{T}: \mathcal{S}\times\mathcal{A}\rightarrow\mathcal{S}$, and reward function $\mathcal{R}:\mathcal{S}\times\mathcal{A}\rightarrow\left[0,1\right]$.
In our LM-based agent scenario, $\mathcal{Q}, \mathcal{A}, \mathcal{O}$ are subsets of language space.

The interaction process between the agent and the environment is described as follows. Given a planning instruction $q_p\in \mathcal{Q}$ that prompts the agent to plan for a task, the agent with parameter $\theta$ generates the first action $a_1\sim \pi_\theta(\cdot|q_p) \in\mathcal{A}$ according to its policy $\pi_\theta$. Each action at step $t$ induces a transformation in the latent state space $s_t \in \mathcal{S}$.
And the agent would face a new observation $o_t\in\mathcal{O}$. Then the agent would incorporate task instruction $q_p$ and interaction trajectories $j_{t} = (a_1,o_1,\dots,a_{t},o_{t})$ to generate the next action $a_{t+1} \sim \pi_\theta(\cdot|q_p,j_{t})$. The interaction loop repeats until the agent assumes the task is finished or the number of steps exceeds the maximum steps.
\vspace{-0.5em}

\subsection{Exploration-based Error Correction Learning}
\label{sec:E2CL}

Our E$^2$CL framework consists of three phases of learning and exploration within the environment: \textit{pre-tuning}, \textit{exploration}, and \textit{training}. In the pre-tuning phase, the agent is equipped with basic planning ability before exploration. Then, the agent collects exploration experiences in the environment via two complementary schemes, as shown in \Cref{fig:framework}. Following this, in the training phase, the agent is trained to align with the environmental knowledge from the collected experiences. With this alignment, the agent is expected to provide feedback by itself and correct its own errors.

\paragraph{Pre-tuning Phase}
To serve as the foundation for environmental exploration, we aim to empower LM-based embodied agents with basic planning capabilities. Given a dataset $\mathcal{J}=\{(q^i_p,j^i_{n_i})\}^{|\mathcal{J}|}$ with $|\mathcal{J}|$ task instructions and expert trajectories where each trajectory has $n_i$ steps, we first construct a planning dataset $D_p$ by slicing each trajectory into sub-trajectories of varying lengths from $1$ to $n_i$. Formally, the planning dataset $D_p$ is defined as: 
\begin{align}
\small
    D_p = \bigcup_{i=1}^{|\mathcal{J}|} \bigcup_{t=1}^{n_i} \left\{(q^i_p, j^i_t) \mid j^i_t \subseteq j^i_{n_i} \land (q^i_p,j^i_{n_i}) \in \mathcal{J} \right\}
\end{align}

Notably, we sample a subset of $D_p$, denoted as $D_{p'}$, for pre-tuning to avoid overfitting to expert trajectories and maintain exploration diversity.
Then, we fine-tune the LM-based agent by minimizing negative log-likelihood loss:
\vspace{-0.6em}
\begin{align}
    \mathcal{L}(\theta) &= \mathbb{E}_{\sim \mathcal{D}_{p'}} \left[ -\log \pi_\theta\left(a_t \mid (q_p, j_{t-1})\right) \right].
\end{align}
where $a_t$ denotes an action consisting of multiple tokens. Therefore, when calculating the loss, it effectively becomes an auto-regressive loss over a sequence of tokens, following previous practices. This approach is consistently applied in the latter stages of training as well.

\paragraph{Exploration Phase}

Intuitively, to gather diverse experiences that fully cover environmental knowledge, we can simply let the agent freely execute its predicted plans and collect the trajectories. 
However, when utilizing these trajectories, we need to correct the errors made by the agent.
Since these trajectories are newly generated, we do not have the correct action data corresponding to the errors. Although one can use a more powerful LLM to correct these errors automatically, the quality of the generated data is inevitably lower compared to expert data.
To balance data diversity and quality, we propose a limited exploration scheme guided by expert trajectories, referred to as \emph{teacher-guided exploration} (TGE). 
Correspondingly, we call the aforementioned free exploration scheme \emph{teacher-free exploration} (TFE).
These two schemes complement each other and enhance the diversity and quality of the collected experiences.

Specifically, for each expert's sub-trajectory $(q_p, j_{t})\in D_p$, the agent conducts TGE by executing the action $\hat{a}_t \sim \pi_\theta(\cdot|q_p, j_{t-1})$.
The environment then provides feedback $f_t$ indicating the executability of this action.
Since the agent only performs one step of exploration under the guidance of the expert, we can naturally use $a_t$ as the ground truth action for that step.
After traversing all the expert trajectories, we obtain the feedback dataset $D_f^{\text{TGE}}$ consisting of samples in the form of $(q_f, j_{t-1}, \hat{a}_t, f_t)$, and the correction dataset $D_c^{\text{TGE}}$ consisting of samples in the form of $(q_c, j_{t-1}, \hat{a}_t, f_t, a_t)$ where $\hat{a}_t\neq a_t$, $q_f$ is the instruction that prompts the model to generate feedback, and $q_c$ is the instruction that prompts the model to correct errors. Please refer to~\Cref{data_template} for templates of these samples.

During TFE, the agent iterates through each task instruction $q_p\sim \mathcal{Q}$ and accrodingly obtain trajectories $j_t=(\hat{a}_1,\hat{o}_1,\dots,\hat{a}_{t},\hat{o}_{t})$. Similar to TGE, whenever the agent predicts a non-executable $\hat{a}_{t}$, the environment provides feedback $f_t$ indicating why this action is non-executable. To obtain the executable action at this step without manual intervention, we leverage an LLM with powerful reasoning ability (e.g., GPT-4o) to automatically correct the action, yielding $a_t$. Considering the LLM may not always provide perfect corrections due to a lack of environment alignment, we further filter its predictions to ensure the corrections are executable. Specifically, the corrected action $a_t$ predicted by LLM would replace original error action $\hat{a}_{t}$ and be executed in the environment. If the corrected action $a_t$ is successful executable in the environment, we would collect these correction data into $D_c^{\text{TFE}}$. As a result, we obtain the feedback dataset $D_f^{\text{TFE}}$ and $D_c^{\text{TFE}}$, each in the same form as the samples in $D_f^{\text{TGE}}$ and $D_c^{\text{TGE}}$, respectively.

\begin{algorithm}[t!]
   \caption{Speculative Inference}
   \label{alg:SI}

\KwIn{$\pi_\theta$: policy of the embodied agent, ES: environment simulator}
\KwOut{R: execution result for each task}

\While{Step length less than threshold}{
    Generate initial action $\hat{a}_t$ \\
    \If{The task is finished}{Iteration stops}
    Generate feedback $\hat{f}_t$ for $\hat{a}_t$ \\
    \If{$\hat{a}_t$ is non-executable}{
    Generate correction action $\hat{a}_c$\\
    Generate feedback $\hat{f}_t$ for $\hat{a}_c$ \\
    \If{$\hat{a}_c$ is executable}{$\hat{a}_c$ executed in ES}
    }
    \Else {$\hat{a}_t$ executed in ES}
    Execution information recorded into R \\
    Renew to next time step
}

\Return{R}
\end{algorithm}

\paragraph{Training Phase}

After the above two phases, we obtain the planning dataset $D_p$, the feedback dataset $D_f=D_f^{\text{TGE}}\bigcup D_f^{\text{TFE}}$, and the correction dataset $D_c=D_c^{\text{TGE}}\bigcup D_c^{\text{TFE}}$. Next, we train the agent to align with the environmental knowledge gathered from these datasets and to develop the ability to provide feedback and correct its own errors.
This is achieved by fine-tuning the agent to minimize the following losses:

\vspace{-0.7em}
\begin{align}
    &\mathcal{L}_p(\theta) = \mathbb{E}_{\sim \mathcal{D}_p} \left[ -\log \pi_\theta(a_t \mid q_p, j_{t-1}) \right], \nonumber\\
    &\mathcal{L}_f(\theta) = \mathbb{E}_{\sim \mathcal{D}_f} \left[ -\log \pi_\theta({f}_t \mid q_f, j_{t-1}, \hat{a}_t) \right], \nonumber\\
    &\mathcal{L}_c(\theta) = \mathbb{E}_{\sim \mathcal{D}_c} \left[ -\log \pi_\theta(a_t \mid q_c, j_{t-1}, \hat{a}_t, {f}_t) \right], \nonumber\\
    &\mathcal{L}_{total}(\theta) = \mathcal{L}_p(\theta) + \mathcal{L}_f(\theta) + \mathcal{L}_c(\theta).
\end{align}

We refer the reader to the pseudo-code of the overall E$^2$CL process in Appendix \ref{Pseudocode}.

\subsection{Speculative Inference}

To utilize the learned abilities in the training phase, we propose \textit{speculative inference} algorithm, a process of inferring errors that may occur ahead of execution and correcting the errors by itself. Based on self-produced feedback, the agent is desired to reduce execution errors and generate correct actions.

To be more precise, when given each test task instruction $q_p$, the agent initially predicts an action $\hat{a}_t \sim \pi_\theta(\cdot|q_p, j_{t-1})$. However, this action $\hat{a}_t$ will not be executed immediately. The agent will ``reflect'' itself and generate an environment feedback $\hat{f}_t \sim \pi_\theta(\cdot|q_f, j_{t-1}, \hat{a}_t)$. If the agent believes the initial action $\hat{a}_t$ is executable, then this action will be executed. Otherwise, the agent will correct this action $\hat{a}$ and predict a new action $\hat{a}_c \sim \pi_\theta (\cdot|q_c, j_{t-1}, \hat{a}_t, \hat{f}_t)$. Once the corrected action $\hat{a}_c$ passes its own check, this action will be executed at this step and concatenated into trajectory $j_{t-1}$. The above process is iterated until the agent assumes the task is finished or the total steps exceed the maximum threshold. 
The process of speculative inference for each test task is shown in ~\Cref{alg:SI}.

\section{Experiments}

\subsection{Experimental Settings}

\paragraph{Embodied Environment \& Tasks}

In this work, we aim to bridge the misalignment gap between LM-based agents and environmental physical constraints, and our method focuses on learning to correct erroneous actions. This implies that providing environmental feedback regarding detailed error information is a necessity. 
To the best of our knowledge, \textbf{VirtualHome}~\citep{puig2018virtualhome}, which focuses on performing typical household tasks, is the most suitable simulated environment for exploration. In VirtualHome, the environmental feedback is in the form of a textual message regarding the physical environmental constraints. If an agent's action is executable, the environment will return a short message of ``True'', otherwise, the environment will return an error message indicating why the action is not executable. For example, one of the typical error types for such feedback is called ``Missing Object'', which arises when the agent is not holding the necessary object to complete an action. More error types of environmental feedback are illustrated in Appendix~\ref{sec:error_type}.

We use the predefined tasks from ActivityPrograms ~\citep{puig2018virtualhome} knowledge base for the experiment. It contains 292 unique high-level household tasks, with 1374 unique action plans and 6201 unique environmental settings in total extracted from VirtualHome. After filtering low-quality tasks, we conduct experiments on a total of 285 tasks. They are randomly divided into a training set of 235 tasks and a test set of 50 tasks. 
We select 50 tasks from the training set as seen tasks, while the 50 tasks in the test set as unseen tasks. We evaluate the method on both seen tasks and unseen tasks. 

Note that during the experiment, we are able to access environmental feedback from the environment simulator. In the testing process, to align with real-world conditions, we do not expect the agents to access such environmental feedback. 
We refer to Appendix \ref{env_illustra} for more details of the environment simulator.

\begin{table*}[t]
\centering
\resizebox{0.95\textwidth}{!}{
\begin{tabular}{llcccccc}
\toprule
\multirow{2}{*}{\textbf{Type}} & \multirow{2}{*}{\textbf{Method}} & \multicolumn{3}{c}{\textbf{Seen Tasks}} & \multicolumn{3}{c}{\textbf{Unseen Tasks}} \\
\cmidrule(lr){3-5} \cmidrule(lr){6-8}
 &  & \textbf{Exec.} & \textbf{AR} & \textbf{LCS} & \textbf{Exec.} & \textbf{AR} & \textbf{LCS} \\
\midrule
Prompting-based & Language-planner~\citep{huang2022language} 
 & 0.18 & 0.62 & 0.19 & 0.16 & 0.58 & 0.16 \\
\hline
\multirow{6}{*}{Tuning-based} 
 & BC & 0.70 & 0.90 & 0.67 & 0.35 & 0.80 & 0.38 \\
 & BC+PPO & 0.70 & 0.91 & 0.68 & 0.37 & 0.81 & 0.38 \\
 & LWM~\citep{xiang2024language} & 0.69 & 0.89 & 0.67 & 0.34 & 0.79 & 0.40 \\
 & Plasma~\citep{Brahman2023PlaSma} & 0.71 & 0.91 & 0.67 & 0.41 & 0.82 & 0.40 \\
 & Lema~\citep{an2023learning} & 0.70 & 0.91 & 0.68 & 0.42 & 0.82 & 0.40 \\
 & NAT~\citep{wang2024learning} & 0.67 & 0.89 & 0.75 & 0.49 & 0.83 & 0.44 \\
\cmidrule{2-8}
 & \textbf{E$^2$CL (Ours)} & \textbf{0.79} & \textbf{0.94} & \textbf{0.78} & \textbf{0.57} & \textbf{0.85} & \textbf{0.46} \\
\bottomrule
\end{tabular}
}
\caption{Comparisons between our method and other baselines on \textit{seen} and \textit{unseen} tasks.}
\label{tab:combined}
\end{table*}

\paragraph{Baselines} 
We compare our method with both prompting-based methods and other tuning-based baseline methods. Similar to our approach, tuning-based methods achieve alignment between the embodied agent and the environment via model fine-tuning.
(1) \textbf{Language-planner}~\citep{huang2022language} aims to inject environment knowledge into prompt and prompted Large Language Models to output action. To better represent prompting-based methods, we use the most powerful LLM, GPT-4o, as the foundation model for our baseline method.
(2) We perform \textbf{Behavior Cloning (BC)} on expert planning data~\citep{chen2023fireact,zeng2023agenttuning}, which is the same method used in the pre-tuning phase of our methods and other baselines.
(3) We conduct \textbf{Proximal Policy Optimization (PPO)}~\citep{schulman2017proximal} after BC. Similar to VirtualHome~\citep{puig2018virtualhome}, we utilize LCS as the reward for RL training.
(4) \textbf{LWM}~\citep{xiang2024language} employs an embodied agent to interact with the environment and collect a large amount of environmental knowledge data to fine-tune the model.
(5) \textbf{Plasma}~\citep{Brahman2023PlaSma} leverages ChatGPT to generate multi-task planning-related data for model training.
(6) \textbf{Lema}~\citep{an2023learning} enhances the agent's reasoning capabilities by providing error-correction data pairs during model fine-tuning.
(7) \textbf{NAT}~\citep{wang2024learning} implements a negative-aware training approach, enabling LM-based agents to effectively learn from both positive and negative examples.

\paragraph{Evaluation Metrics}

Following previous studies~\citep{puig2018virtualhome,raman2022cape}, we evaluate our action plans across three metrics: \textbf{executability (Exec.)}, \textbf{affordance rate (AR)}, and \textbf{longest common sequence (LCS)}.
Executability measures whether an action plan can be correctly parsed and satisfies the common-sense constraints of the environment. Specifically, the parsed action must contain only allowable action, and the objects must be in the environment. Moreover, the action must satisfy the pre-conditions (e.g., the embodied agent cannot send email before walking to the computer) and post-conditions (e.g., the state of TV changes from closed to open after the agent opens it). Similarly to executablility, affordance rate measures the average percentage of all plan steps that are executable, in cases where the entire plan is not executable. However, executability and affordance rate only can reflect whether the agent could compliant with environment physical constraints, but they cannot reflect whether the plan is correct. LCS calculates the length of the longest common subsequence between generated plans and the ground truth plans, normalized by the maximum length of the two. It can reflect the correctness of the plans generated by the agents.

\subsection{Experimental Results}

Our experimental results are shown in ~\Cref{tab:combined}. As can be seen, the prompting-based method significantly lags behind all tuning-based methods in different metrics.
Despite this contradicts with the experience that LLMs exhibit exceptional general reasoning capabilities, we observe that the actions generated by prompt-based methods, while seemingly reasonable, fail to comply with the physical constraints of the environment often.

Regarding tuning-based baseline methods, our E$^{2}$CL method demonstrates significant improvements over BC in both seen and unseen tasks. 
Even with PPO applied on top of BC, the performance remains weak. This is likely because the action space is too large and the rewards are sparse, making it difficult to optimize the model in such an embodied environment.
Moreover, LWM and Plasma, which are also fed by expert planning data and can be seen as augmented versions of BC, only show a marginal increase in performance.
Compared to these BC-based methods, the method utilizing failure data, i.e., Lema and NAT, demonstrates better performance. Taking a step further, we evolve this idea by training the agent to develop self-feedback and self-correction capabilities through its failure experiences. The results show that our method increases executability-related metrics by up to 15\% and LCS by up to 10\% compared with Lema and NAT. This demonstrates that these two capabilities effectively enable the agent to align with the environment for task-solving.

\subsection{Ablation Study on Training Data}

\begin{table}[t]
\centering
\resizebox{0.98\linewidth}{!}{
\begin{tabular}{lcccccc}
\toprule
\textbf{Method} & & & &\ \textbf{Exec.} & \ \textbf{AR} & \ \textbf{LCS}   \\
\midrule
Ours                   & & &  &  \textbf{0.57}    &  \textbf{0.85}   &  \textbf{0.46}        \\
- w/o $D_c$ and $D_f$   & & &  &  0.35    &  0.80   &  0.38       \\
- w/o $D_f$             & & &  &  0.44    &  0.82   &  0.39        \\
- w/o $D_c$             & & &  &  0.41    &  0.82   &  0.39        \\
\bottomrule
\end{tabular}
}
\caption{Task-solving performance of the agent on unseen tasks when trained with ablated data.}

\label{tab:ablation}
\end{table}

In this section, we explore the impact of the collected training data, i.e., feedback data $D_f$ and correction data $D_c$, on overall performance by ablating them in the training. During the inference phase, we employ speculative inference for all settings to ensure consistency.

As shown in~\Cref{tab:ablation}, we observe that both $D_f$ and $D_c$ are each beneficial for the agent, but lag behind the combination of them. 
We hypothesize that the improvement observed when training with $D_c$ is primarily due to the enhanced self-correction capability of the agent. However, the limited ability to generate high-quality action feedback hampers the effectiveness of self-correction during speculative inference, as demonstrated in Section \ref{section:si}.
Compared to the agent training without both $D_f$ and $D_c$, training with $D_p$ and $D_f$ improves the performance by training to predict environmental feedback, which explicitly aligns with environmental knowledge. However, the weak self-correction capability of the agent constrains the agent from generating executable and correct action in speculative inference, which is demonstrated in ~\Cref{section:sc}.
In our method, we integrate both types of data, enabling our agent to generate higher-quality action feedback and exhibit stronger self-correction abilities. This results in a substantial performance boost compared to other ablation settings.

\subsection{Analysis on Different Size of the Model}

\begin{figure}
  \includegraphics[width=\columnwidth]{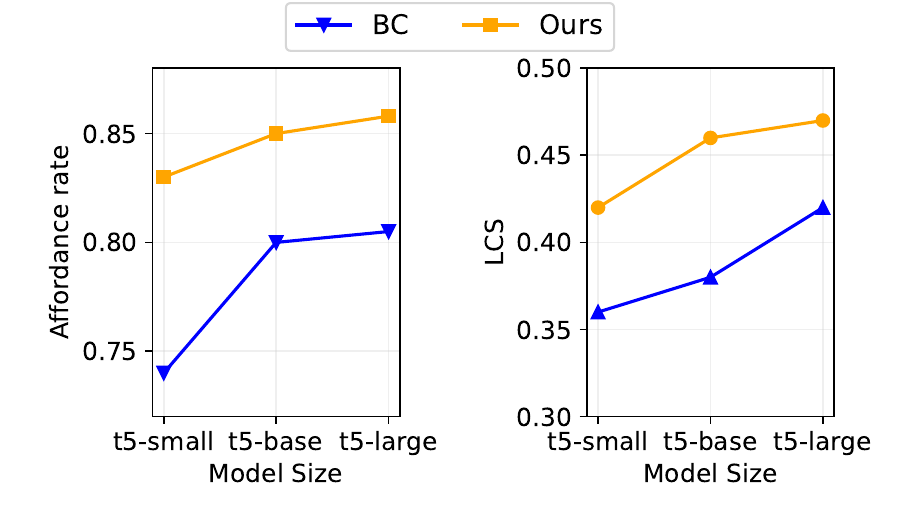}
  \caption{Task-solving performance of the agent on unseen tasks based on different sizes of LM and different training methods.}
  \label{fig:LLM_size}
\end{figure}

To investigate the impact of model size on performance, we train models of different sizes using both BC and our method, and evaluated them on unseen tasks.
Our results are shown in Figure \ref{fig:LLM_size}, where larger models perform relatively better across all aspects, indicating that model scale significantly impacts performance.
Moreover, it can also be observed that our method outperforms BC in both affordance rate and LCS across models with different parameter sizes, which demonstrates that our method consistently provides superior performance regardless of model size.
Notably, when using our method, smaller models achieve performance surpassing larger models using BC across all metrics.
This finding suggests that our method is able to release the potential of small language models and lays the foundation for building agents that work on edge devices in the future.

\subsection{Evaluation on Self-Correction Ability}
\label{section:sc}

We further evaluate the self-correction capability of our constructed agent. 
We conduct two different experiment settings to validate the performance of the agent. For seen tasks, we randomly select 100 samples from correction data. For unseen tasks, we collect 100 correction data samples in a similar process to TGE.
For comparison, we also evaluate the prompting-based agent and the agent trained by BC. Since these two agents have both undergone general instruction tuning, we instruct them to conduct self-correction off the shelf.

\begin{figure}
  \includegraphics[width=\columnwidth]{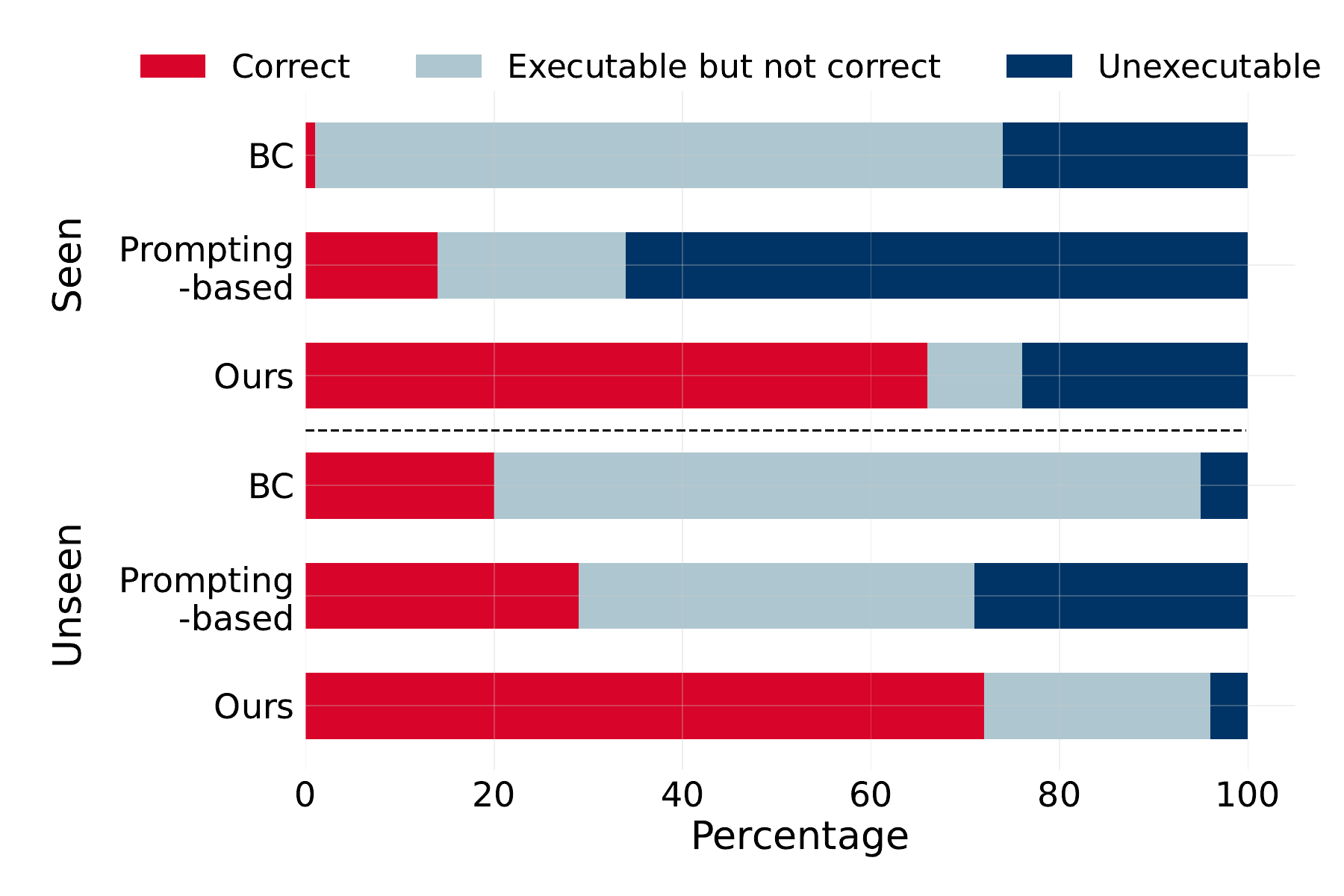}

  \caption{Comparison of self-correction capability between our method and other baseline methods.}
  \label{fig:percen}
\end{figure}

As shown in ~\Cref{fig:percen}, our method generates correct corrected actions far more frequently than BC and prompting-based methods in both seen tasks and unseen tasks, which demonstrates our agent's strong self-correction capability.
The powerful self-correction capability reflects our agent can truly align with the environment and generate correct corrective actions that do not violate physical constraints. 
Furthermore, we can observe from~\Cref{fig:percen} that our agent generates correct actions at a high proportion in both seen and unseen tasks. This ensures a reliable self-correction process in speculative inference.

\begin{table}[t]
\centering
\tabcolsep=3pt
\resizebox{\linewidth}{!}{
\begin{tabular}{lccccc}
\toprule
\textbf{Method} & \ \textbf{w/ $\bm{D_f}$\&$\bm{D_c}$?} & \ \textbf{w/ SI?}  & \ \textbf{Exec.} & \ \textbf{AR} & \ \textbf{LCS} \\
\midrule
   & \xmark   & \xmark &  0.35    &  0.80  & 0.38       \\
Ours   & \cmark  & \xmark &  0.47    &  0.82  & 0.46    \\
   & \cmark & \cmark  &  \textbf{0.57}    &  \textbf{0.85}  & \textbf{0.46}  \\
\bottomrule
\end{tabular}
}
\caption{Performance of the agent regarding speculative inference (SI).}
\label{tab:self-correction}
\end{table}

\subsection{Analysis on Speculative Inference}
\label{section:si}

To analyze the contribution of speculative inference to overall performance, as well as to explore the quality and effectiveness of self-generated feedback, we conduct the following analysis.

Firstly, as shown in ~\Cref{tab:self-correction}, we conduct three kinds of experiment settings and test their performance on unseen tasks. Employing speculative inference significantly improves the agent's executability and affordance rate. This shows that speculative inference effectively reduces errors during execution, which demonstrates the effectiveness of the design. Moreover, LCS has not changed regardless of using speculative inference. This indicates that speculative inference contributes to the performance gain mainly by generating more executable actions, instead of recovering the expert trajectories in the training data.

\begin{figure}[t]
  \includegraphics[width=\columnwidth]{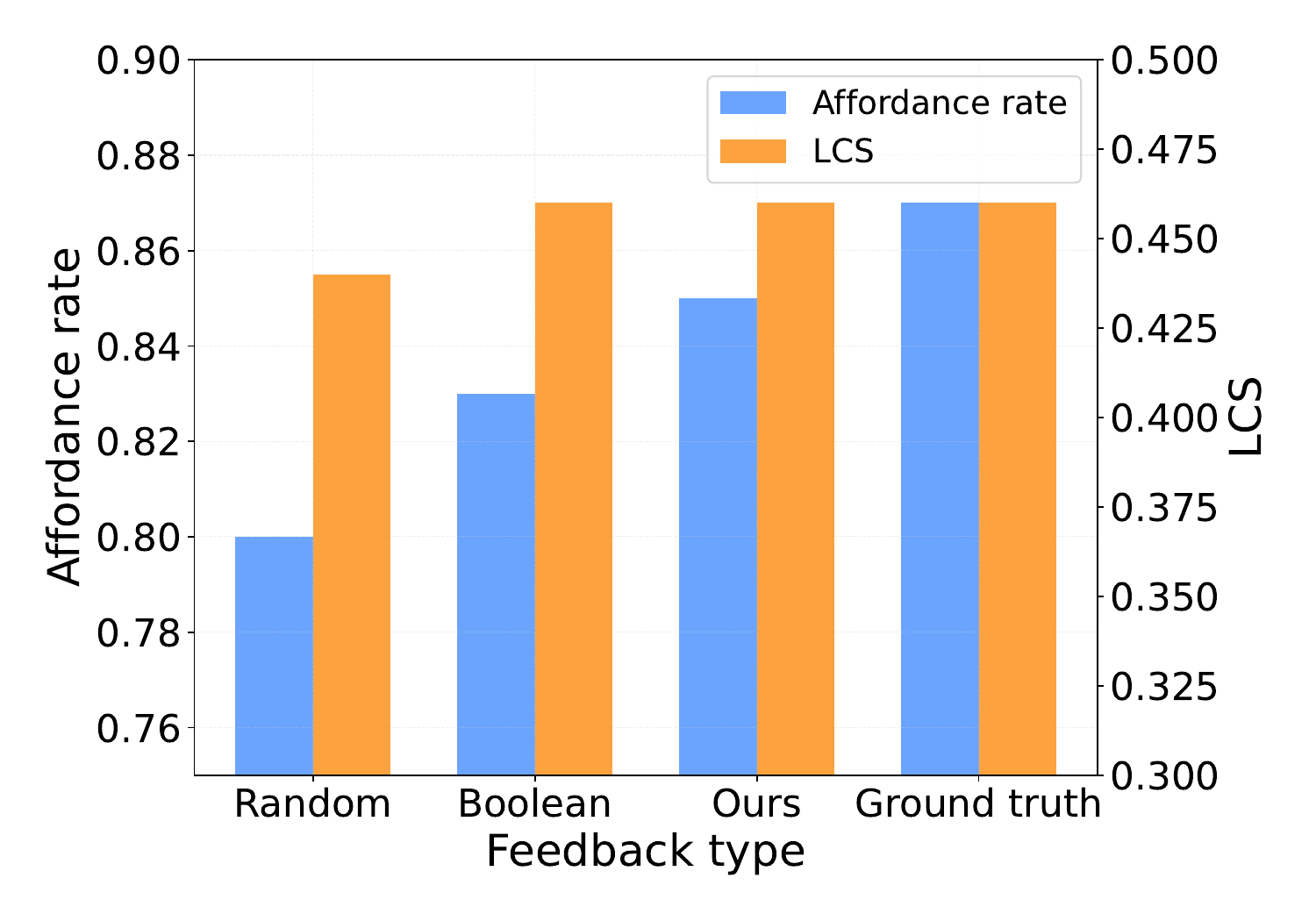}
  \caption{Task-solving performance of the agent on unseen tasks when fed with different types of feedback. \textit{Random}: Randomly select a feedback type from all available feedback. \textit{Boolean}: a ground truth boolean signal, indicating whether the initial action is executable or not. \textit{Ours}: The self-generated feedback on the initial action used in our method. \textit{Ground truth}: The ground truth feedback from the environment.}
  \label{fig:experiment setting}
\end{figure}

Next, we provide our agent with self-generated feedback as well as three other types of feedback, and test its performance on unseen tasks. 
As shown in ~\Cref{fig:experiment setting}, given random feedback to the agent, the agent performs worst in both affordance rate and LCS, which underscores the importance of high-quality feedback.
When fed with self-generated feedback, the agent performs better than that of using random feedback and boolean executability signals, while slightly worse than that of using ground truth. This suggests that our method enables the agents to generate feedback with good quality.
Overall, we can observe that feedback with better qualities yields a better performance, which demonstrates that the speculative inference process faithfully relies on high-quality feedback.

\subsection{Error Analysis}

We also perform an error analysis to identify the aspects where the agent constructed using our method outperforms BC. There are a total of eight types of errors, which can be further classified into grounding errors (object availability) and execution-related errors (others). The detailed demonstration can be found in the Appendix \ref{sec:error_type}. 
As shown in Figure \ref{fig:error analysis}, we observe that all error types decreased by more than 24\%, with \textit{Over occupied} error showing the highest reduction rate of 94.4\%. This demonstrates the effectiveness of our method in reducing various types of errors, highlighting its comprehensiveness.
For the two most frequent types of execution-related errors, \textit{unflipped boolean state} and \textit{agent proximity}, our method achieves a reduction in error count by over 37\% compared to BC, thereby demonstrating its effectiveness.
Although our method primarily aims to avoid execution errors related to physical constraint and does not specifically target grounding errors such as \textit{object availability}, the fact that it still reduces this type of error demonstrates the generalizability of our method.

\begin{figure}
  \includegraphics[width=\columnwidth]{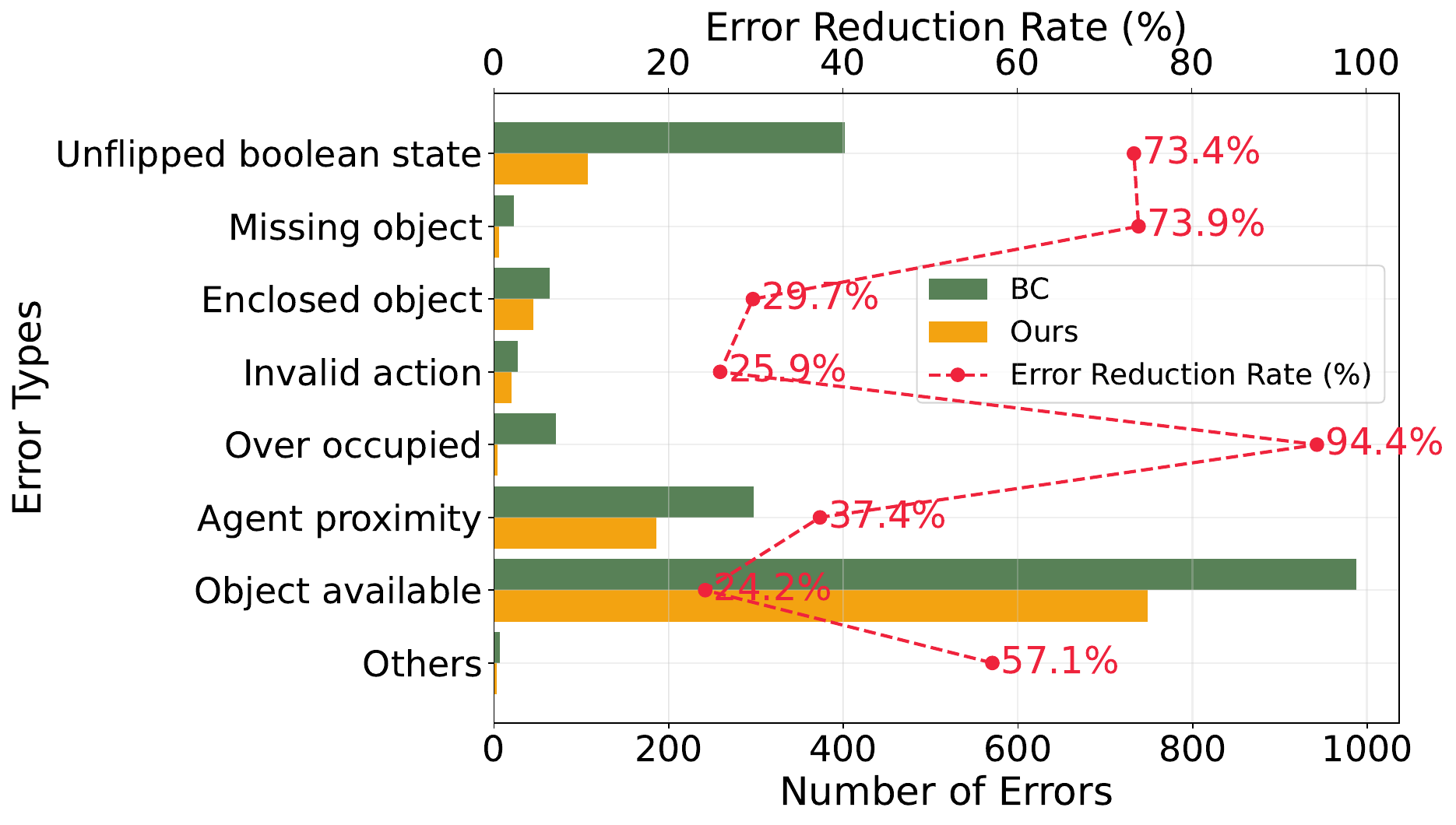}
  \captionsetup{hypcap=false} 
  \captionof{figure}{Error statistics of our method and BC when testing on unseen tasks.}
  \label{fig:error analysis}
  \captionsetup{hypcap=true}  
\end{figure}

\section{Related Work}

\paragraph{LM-based Agent} 

Nowadays, due to the increasingly powerful generalization capabilities of language models, they are often regarded as the policy function of agents to plan their behavior~\citep{tan2024true, carta2023grounding}. However, one issue is that there may be a misalignment between the knowledge in the environment and the internal knowledge of the model. Consequently, a significant amount of work aims to ground the language model to the environment~\citep{brohan2023can, fu2024scene, song2023llm}. Some studies harness the immense capabilities of large language models and employ intricate prompts or integrate specifically designed modules~\citep{huang2022language,huang2024grounded,raman2022cape,singh2023progprompt,wang2305voyager,guan2023leveraging}. However, LLM-based agents would cost heavily and are not suitable for offline scenarios. Some line of work deploys language model as decision-making agents to align with embodied environments via reinforcement learning~\citep{tan2024true, carta2023grounding}. This type of approach tends to have low learning efficiency in embodied environments with large action spaces. In addition, similar to our approach, other research efforts have proposed frameworks where the agent first explores the environment and subsequently utilizes the exploration experience for learning~\citep{li2022pre,xiang2024language}. These approaches often overly focus on the agent and lack comprehensive environmental feedback modeling, making it difficult to avoid execution errors.

\paragraph{Learning from Failure}

After exploration, the agent would encounter failure in the past experience, which is assumed as negative samples. The topic of learning from negative samples has increasingly gained attention as an alternative approach to learning solely from positive samples. Traditionally, some studies aim to decrease the probability of negative samples while increasing the probability of positive samples in order to achieve better performance~\citep{wang2024aligning,zheng2023click,liu2022brio}. Additionally, some works construct correcting dataset and tuning language models on these data~\citep{an2023learning,wang2024learning,bai2022constitutional}. Besides, there are other efforts aimed at leveraging the comprehension abilities of language models to widen the gap between positive and negative samples~\citep{liu2023chain,zhang2023wisdom, tong2024can}. In our work, we similarly leverage the inherent understanding capabilities of language models and enhance the embodied agent's learning from environmental feedback regarding exploration errors, as well as its ability to self-correct.

\section{Conclusion}

In this work, we aim to align the embodied agent with the environment to enhance its task-solving performance. 
Firstly, we present E$^2$CL, a novel framework that leverages exploration-induced errors and environmental feedback to enhance environment alignment for LM-based agents during \textit{teacher-guided} and \textit{teacher-free} exploration. 
Furthermore, we introduce speculative inference, a process in which the agent utilizes learned abilities for self-feedback and self-correction to reduce execution errors. Extensive experiments show that our method outperforms many existing baseline methods.

\section*{Limitations}

The baseline model for the embodied agent constructed using our method is a text-based model, meaning the agent's observations are input in textual form. However, there is a gap between textual descriptions of real-world visual images and the actual visual information, which cannot fully encapsulate all real-world details. This discrepancy affects the robot agent's ability to ground itself in the environment. In future work, we aim to incorporate visual information directly into the input to better align with real-world scenarios. Additionally, although VirtualHome~\citep{puig2018virtualhome} is a relatively complex environment, we have not conducted experimental validation in other embodied environments or the real world. In the future, we will perform more experiments for validation.

\section*{Ethical Considerations}

This work aims to construct an embodied agent within Virtual Environment. 
The virtual environment setup and related data strictly follow the specifications of VirtualHome~\citep{puig2018virtualhome}. 
We refer to VirtualHome v2.3.0\footnote{\url{https://github.com/xavierpuigf/virtualhome/tree/master}} to conduct out our experiments (MIT license\footnote{\url{https://github.com/xavierpuigf/virtualhome/blob/master/LICENSE}}).
The models, i.e. flan-t5-small, flan-t5-base and flan-t5-large~\citep{chung2024scaling}, we use for fine-tuning are all open-source, and we will strictly follow the protocols for the academic use of these language models (Apache License 2.0\footnote{\url{https://huggingface.co/google/flan-t5-base}}). 
In addition, we partially use AI assistants, such as Copilot and ChatGPT, to help with our coding and writing.


\section*{Acknowledgements} 
This work was supported by the National Natural Science Foundation of China (62076212) and the Research Grants Council of Hong Kong (15209724). The authors would like to thank the anonymous reviewers for their valuable feedback and constructive suggestions.

\bibliography{acl_latex}

\appendix

\newpage 
\section*{Appendix}
\label{sec:appendix}

\section{Data}
\label{data_template}


\begin{figure}[ht]
  \includegraphics[width=\columnwidth]{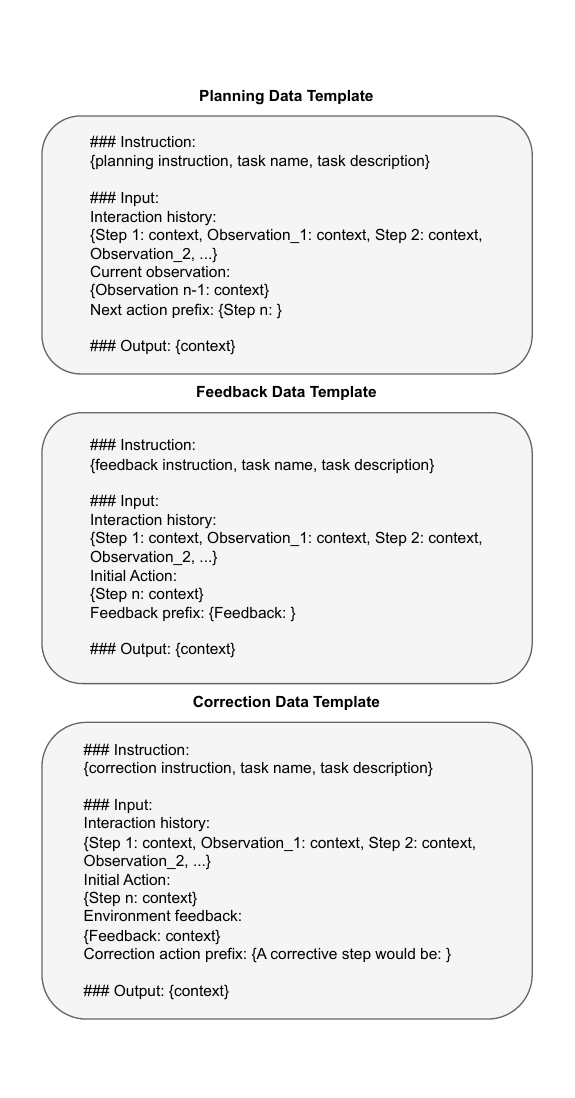}
  \caption{Data templates of planning data, feedback data and correction data.}
  \label{fig:Data_visualization}
\end{figure}

\begin{figure}[ht]
  \includegraphics[width=\columnwidth]{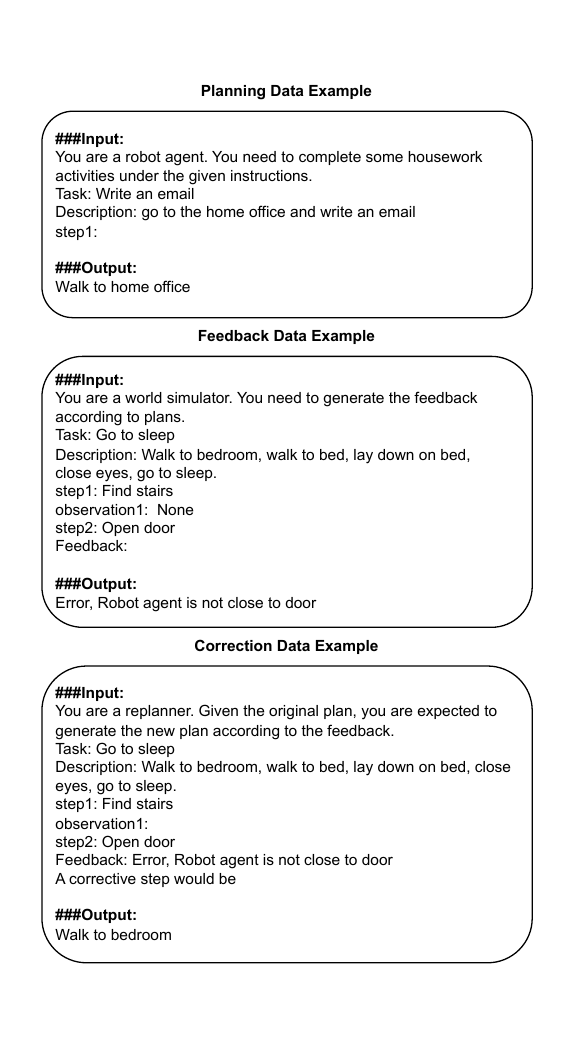}
  \caption{Data examples of planning data, feedback data and correction data.}
  \label{fig:Data_example_visualization}
\end{figure}

\section{Additional Details of VirtualHome Environment}
\label{env_illustra}
VirtualHome provides diverse and customizable household environments that support a wide array of possible interactions in the form of atomic action steps. There are three kinds of action template based on the action type, which are "[Action]", "[Action] <Object> <id>" and "[Action] <Object> <id> <Object> <id>".  Each [Action] refers to one of 42 atomic actions supported in Virtualhome. Full list of atomic actions are shown in ~\Cref{tab:atomic actions}.
In each scene, there are approximately 350 objects with which the embodied agent can interact, each identified by a specific <id>. These objects have properties (e.g., drinkable, eatable) corresponding to their action affordances. Some objects also possess a semantic state, such as heated, washed, or used.

In ActivityPrograms ~\citep{puig2018virtualhome} knowledge base, there are 292 unique high-level household tasks, with 1374 unique action plans and 6201 unique environments in total extracted from VirtualHome, and task and action plan samples manually annotated by Amazon Mechanical Turk workers. 
Each data sample consists of a high-level task, a description of the task, and complete action programs that can be directly executed in the VirtualHome environment.
A piece of data sample is shown in ~\Cref{tab:ds}. 

\begin{table}[t]
\centering
\resizebox{\linewidth}{!}{
\begin{tabular}{cc}
\toprule
\multicolumn{2}{c}{\textbf{Task}: Work} \\
\multicolumn{2}{c}{\textbf{Description}: I walk to the office. I sit down in the chair.} \\
\multicolumn{2}{c}{I turn on the computer. I start typing on the keyboard.} \\
\midrule
\textbf{Natural language} & \textbf{Programs} \\
\hdashline
Walk to home office       & [WALK] <home\_office> (319) \\
Walk to chair             & [WALK] <chair> (356) \\
Find chair                &  [FIND] <chair> (356) \\
Sit on chair  &  [SIT] <chair> (356) \\
Find computer  &  [FIND] <computer> (417) \\
Switch on computer  &  [SWITCHON] <computer> (417) \\
Find keyboard  &  [FIND] <keyboard> (415)  \\
Type on keyboard  &  [TYPE] <keyboard> (415) \\
\bottomrule
\end{tabular}
}
\caption{A piece of data sample from ActivityPrograms knowledge base}
\label{tab:ds}
\end{table}

\section{Additional Analyses of Experiment Results}

\subsection{Illustration of Error Types}
\label{sec:error_type}

During the interaction between the agent and the environment, we collect error feedback from the environment and classify it into eight categories as followings.

\textit{Unflipped Boolean State} error occurs when an action meant to change the state of an object with a Boolean attribute (such as open/closed or on/off) does not achieve the intended effect, like attempting to open an already open door. \textit{Missing Object} error arises when the agent is not holding the necessary object to complete an action, preventing the task's execution. \textit{Enclosed Object} error occurs when the target object is contained within a closed structure, preventing the action from freeing the object for use. \textit{Invalid Action} error occurs when the agent attempts to perform an action on a target object that is not afforded to it, such as trying to pull a ceiling. \textit{Over-occupied Agent} error happens when the agent's hands are occupied or already interacting with objects, leaving it unable to interact with the target object in the current step. \textit{Agent Proximity} errors arise when the agent is not close enough to the target object to perform the action. \textit{Object availability} errors occur when the agent attempts to interact with an object that does not exist in the environment. The remaining errors are categorized as \textit{Others}.

\subsection{Length Analysis}

\begin{figure}
  \includegraphics[width=\columnwidth]{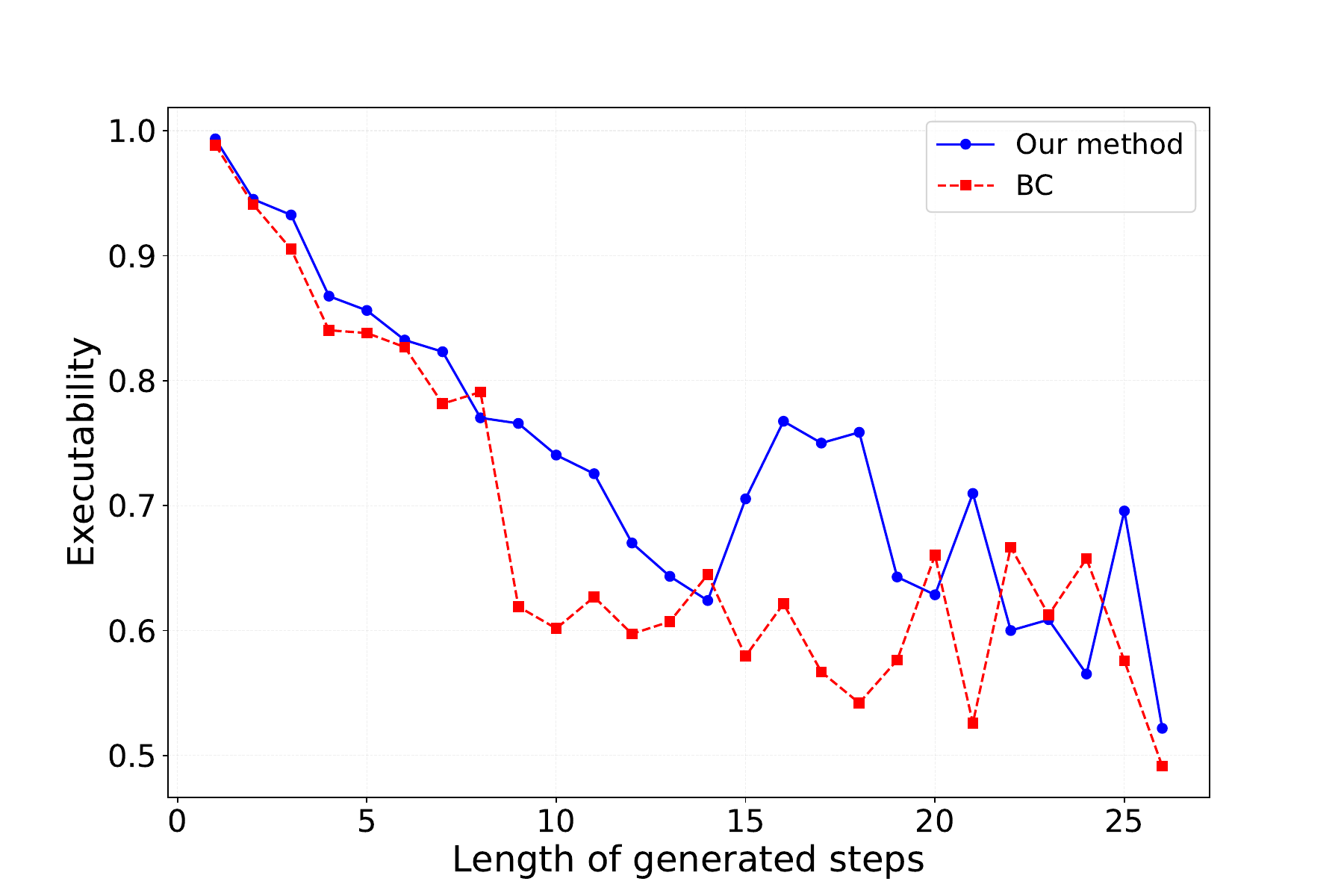}
  \captionsetup{hypcap=false} 
  \captionof{figure}{Comparison of Our method and BC in terms of executability across different tasks (varying lengths of generated steps)}
  \label{fig:prob-len}
  \captionsetup{hypcap=true}  
\end{figure}

Following our common sense, tasks with a greater number of steps are generally considered more challenging for the agent. To evaluate the performance of the agent on tasks of varying difficulty, we collected and analyzed the executability of tasks with different lengths of generated steps between our method and BC. As shown in Figure~\ref{fig:prob-len}, in terms of execution rates for different lengths of generated steps, our method outperforms BC, particularly in tasks with longer steps. This indicates the widespread efficacy of our method.

\subsection{Convergence Speed Analysis}
\label{Convergence_speed}

We report the variations of training losses among our method and two representative baseline methods. The results are shown in the Table~\ref{tab:convergence_speed}. We observe that as the number of epochs increases, the loss for our method decreases more rapidly than that of BC and NAT, ultimately converging at a lower bound. It indicates that our method contributes to the convergence speed during model training.

\begin{table}[t] 
\centering
\tabcolsep=3pt
\resizebox{\linewidth}{!}{
\begin{tabular}{lccccccc}
\toprule
\textbf{Loss} & \ epoch 1 & \ epoch 2  & \ epoch 3 & \ epoch 4 & \ epoch 5  & \ epoch 6 & \ epoch 7 \\
\midrule
\textbf{BC} & 4.4470 & 0.2956 & 0.2631 & 0.1987 & 0.1476 & 0.1075 & 0.0967 \\
\textbf{NAT} & 5.8337 & 0.2164 & 0.1308 & 0.1207 & 0.0758 & 0.0600 & 0.0503 \\
\textbf{Ours} & 5.3267 & 0.1252 & 0.1134 & 0.0849 & 0.0708 & 0.0423 & 0.0354 \\
\bottomrule
\end{tabular}
}
\caption{Comparison of convergence speed between our method and baseline methods}
\label{tab:convergence_speed}
\end{table}

\begin{table}[t] 
\centering
\tabcolsep=3pt
\resizebox{\linewidth}{!}{
\begin{tabular}{ccc}
\toprule
\textbf{Atomic actions} & \  & \ \textbf{VirtualHome accepted format}  \\
\midrule
    CLOSE   & &  [CLOSE] <Object> <id>   \\
    DRINK   & &   [DRINK] <Object> <id>  \\
    FIND   & &   [FIND] <Object> <id>  \\
    WALK   & &   [WALK] <Object> <id>  \\
    GRAB   & &   [GRAB] <Object> <id> \\
    LOOKAT & &   [LOOKAT] <Object> <id> \\
    OPEN & &     [OPEN] <Object> <id> \\
    POINTAT & &   [POINTAT] <Object> <id> \\
    PUTBACK & &   [PUTBACK] <Object> <id> <Object> <id>  \\
    PUTIN & &   [PUTIN] <Object> <id> <Object> <id>  \\
    PUTOBJBACK & &  [PUTOBJBACK] <Object> <id> \\
    RUN & &   [RUN] <Object> <id> \\
    SIT & &   [SIT] <Object> <id> \\
    STANDSUP  & &   [STANDSUP] \\
    SWITCHOFF & &  [SWITCHOFF] <Object> <id> \\
    SWITCHON & &   [SWITCHON] <Object> <id> \\
    TOUCH & &   [TOUCH] <Object> <id> \\
    TURNTO & &   [TURNTO] <Object> <id> \\
    WATCH & &   [WATCH] <Object> <id> \\
    WIPE & &   [WIPE] <Object> <id> \\
    PUTON & &   [PUTON] <Object> <id> \\
    PUTOFF & &   [PUTOFF] <Object> <id> \\
    GREET & &   [GREET] <Object> <id> \\
    DROP & &   [DROP] <Object> <id> \\
    READ & &   [READ] <Object> <id> \\
    LIE & &  [LIE] <Object> <id> \\
    POUR & &   [POUR] <Object> <id> <Object> <id>  \\
    TYPE & &  [TYPE] <Object> <id> \\
    PUSH & &   [PUSH] <Object> <id> \\
    PULL & &   [PULL] <Object> <id> \\
    MOVE & &   [MOVE] <Object> <id> \\
    WASH & &   [WASH] <Object> <id> \\
    RINSE & &   [RINSE] <Object> <id> \\
    SCRUB & &   [SCRUB] <Object> <id> \\
    SQUEEZE & &  [SQUEEZE] <Object> <id> \\
    PLUGIN & &   [PLUGIN] <Object> <id> \\
    PLUGOUT & &  [PLUGOUT] <Object> <id> \\
    CUT & &   [CUT] <Object> <id> \\
    EAT & &   [EAT] <Object> <id> \\
    SLEEP & &   [SLEEP] \\
    WAKEUP & &   [WAKEUP] \\
    RELEASE & &  [RELEASE] <Object> <id> \\
\bottomrule
\end{tabular}
}
\caption{Full list of atomic actions and accepted form of virtualhome environment}
\label{tab:atomic actions}
\end{table}

\section{Additional Implementation Details}
In our work, we primarily fine-tuned three models of different sizes: flan-t5-small with 77 million parameters, flan-t5-base with 248 million parameters, and flan-t5-large with 783 million parameters~\citep{chung2024scaling}. 

All experiments were conducted on eight NVIDIA RTX A6000 GPUs. During the pre-tuning phase, we selected 1000 samples from the expert planning data $D_p$ and trained for one epoch. During the training process, we set the following hyperparameters: a batch size of 30, training for three epochs, and selecting the best-performing checkpoints from these epochs. The learning rate was set to 1e-4. During the inference process, all generation parameters were kept consistent with the default generation parameters of the flan-t5 series models. The results reported in the paper are all averages. All experiments are expected to reproduce in one day.

\begin{algorithm*}[t!]
\DontPrintSemicolon
\KwIn{$\mathcal{D}_{p}$: Expert planning data, $\pi_\theta$: initial robot agent policy, $T_1$: number of epochs in pre-tuning phase, ES: Environment Simulator, $|\mathcal{J}|$: Number of tasks, $n_i$: the step length of task i, M: GPT-4o, $T_2$: number of epochs in training phase.}
\KwOut{Final policy $\pi_\theta$}

\textcolor{dm-blue-500}{\texttt{// Construct a weak robot agent to explore the environment}}\;
Randomly select few planning training data $\mathcal{D}_{p'} \subseteq \mathcal{D}_{p}$\;
\For{$h=1$ \KwTo $T_1$}{
    Optimize $\theta$ on BC objective:     $\mathcal{L}(\theta) = \mathbb{E}_{(q_p,j_{t}) \sim \mathcal{D}_{p'}} \left[ -\log \pi_\theta\left(a_t \mid (q_p, j_{t-1})\right) \right] $ \;
} 
\textcolor{dm-blue-500}{\texttt{// Teacher-guide Exploration}}\;
\For{$i=1$ \KwTo $|\mathcal{J}|$}{
    \For{$t=1$ \KwTo $n_i$}{
        Predicting the action $\hat{a}_t \sim \pi_\theta(q_p,j_{t-1})$ \;
        $\hat{a}_t$ executed in ES and obtain new observation $o_t$, environmental execution feedback $f_t$\;
        \If{($\hat{a}_t \neq a_t$) \& \( \hat{a}_t \text{ is non-executable} \)}
        {
             Correction data sample: ($q_c, j_{t-1},\hat{a}_t, f_t, a_t$) added to $D_c$ \;
             Feedback data sample: ($q_f, j_{t-1}, \hat{a}_t, f_t$) added to $D_f$ \;
            $a_t$ executed in ES and obtain new observation $o_t$ \;
        }
         \If{$\hat{a}_t == a_t$}
         {Feedback data sample: ($q_f, j_{t-1}, a_t, True$) added to $D_f$}
    }
}
\textcolor{dm-blue-500}{\texttt{// Teacher-free Exploration}}\;
\For{$i=1$ \KwTo $|\mathcal{J}|$}{
    \While{the agent assumes the task is not finished}{
         Predicting the action $\hat{a}_t \sim \pi_\theta(q_p,j_{t-1})$ \;
         $\hat{a}_t$ executed in ES and obtain new observation $o_t$, environmental execution feedback $f_t$  \;
         \If{$\hat{a}_t$ is non-executable}
         {
            Gain corrected action $a_t \sim M(q_c, j_{t-1}, \hat{a}_t, f_t)$ \;
            Correction data sample: ($q_c, j_{t-1}, \hat{a}_t, f_t, a_t$) added to $D_c$
         }
         Feedback data sample: ($q_f, j_{t-1}, \hat{a}_t, f_t$) added to $D_f$
    }
}
\textcolor{dm-blue-500}{\texttt{// Learning from Exploration Experience}}\;
\For{$h=1$ \KwTo $T_2$}{
    Optimize $\theta$ on autoregressive objective loss: $\mathcal{L}_{\mathrm{SFT}} (\pi_\theta) = \mathbb{E}_{\sim \mathcal{D}_p} \left[ -\log \pi_\theta(a_t \mid q_p, j_{t-1}) \right] + \mathbb{E}_{\sim \mathcal{D}_f} \left[ -\log \pi_\theta({f}_t \mid q_f, j_{t-1}, \hat{a}_t) \right] + \mathbb{E}_{\sim \mathcal{D}_c} \left[ -\log \pi_\theta(a_t \mid q_c, j_{t-1}, \hat{a}_t, {f}_t) \right] $\;
}
\Return{$\pi_\theta$}
\caption{\textbf{Exploration-based Error Correction Learning}}
\label{algo:ours}
\end{algorithm*}

\section{Pseudocode}
\label{Pseudocode}

This section presents the pseudocode of E$^2$CL in ~\Cref{algo:ours}. A detailed discussion of the method is given in ~\Cref{sec:E2CL}.

\end{document}